\documentclass[11pt,a4paper]{article}
\usepackage[hyperref]{emnlp2020}
\usepackage{times}
\usepackage{latexsym}
\usepackage{hyperref}
\usepackage{multirow}
\usepackage{wasysym}
\usepackage{CJKutf8}
\usepackage[russian,english]{babel}
\usepackage{booktabs}

\usepackage{microtype}

\aclfinalcopy 
\def\aclpaperid{1961} 

\usepackage{todonotes}

\title{On Romanization for Model Transfer Between Scripts\\ in Neural Machine Translation}

\author{Chantal Amrhein$^1$ \and Rico Sennrich$^{1,2}$\\
  $^1$Department of Computational Linguistics, University of Zurich\\
  $^2$School of Informatics, University of Edinburgh \\ \medskip
  \texttt{\{amrhein,sennrich\}@cl.uzh.ch}}

\date{}

\begin{document}
\maketitle
\begin{abstract}
Transfer learning is a popular strategy to improve the quality of low-resource machine translation. For an optimal transfer of the embedding layer, the child and parent model should share a substantial part of the vocabulary.  
This is not the case when transferring to languages with a different script. We explore the benefit of romanization in this scenario. Our results show that romanization entails information loss and is thus not always superior to simpler vocabulary transfer methods, but can improve the transfer between related languages with different scripts. We compare two romanization tools and find that they exhibit different degrees of information loss, which affects translation quality.
Finally, we extend romanization to the target side, showing that this can be a successful strategy when coupled with a simple deromanization model.
\end{abstract}

\section{Introduction}
Neural Machine Translation (NMT) has opened up new opportunities in transfer
learning from high-resource to low-resource language pairs \citep{zoph-etal-2016-transfer, kocmi-bojar-2018-trivial, lakew2018transfer}.
While transfer learning has shown great promise, the transfer between
languages with different scripts brings additional challenges. For a successful transfer of the embedding layer, both the parent and the child model should use the same or a partially overlapping vocabulary \citep{aji-etal-2020-neural}. It is common to merge the two vocabularies by aligning identical subwords and randomly assigning the remaining subwords from the child vocabulary to positions in the parent vocabulary \citep{lakew2018transfer, lakew2019adapting, kocmi-bojar-2020-efficiently}. 

This works well for transfer between languages that use the same script, but if the child language is written in an unseen script, most vocabulary positions are replaced by random subwords. This significantly reduces the transfer from the embedding layer. \citet{gheini2019universal} argue that romanization can improve transfer to languages with unseen scripts. However, romanization can also introduce information loss that might hurt translation quality. In our work, we study the usefulness of romanization for transfer from many-to-many multilingual MT models to low-resource languages with different scripts. Our contributions are the following: 
\begin{itemize}
    \item[-] We show that romanized MT is not generally optimal, but can improve transfer between related languages that use different scripts.
    \item[-] We study information loss from different romanization tools and its effect on MT quality.
    \item[-] We demonstrate that romanization on the target side can also be effective when combined with a learned deromanization model.
\end{itemize}

\section{Related Work}
Initial work on transfer learning for NMT has assumed that the child language is known in advance and that the parent and child model can use a shared vocabulary \citep{nguyen-chiang-2017-transfer, kocmi-bojar-2018-trivial}. \citet{lakew2018transfer} argue that this is not feasible in most real-life scenarios and propose using a dynamic vocabulary. Most studies have since opted to replace unused parts of the parent vocabulary with unseen subwords from the child vocabulary \citep{lakew2019adapting, kocmi-bojar-2020-efficiently}; others use various methods to align embedding spaces \citep{gu-etal-2018-universal, kim-etal-2019-effective}. Recently, \citet{aji-etal-2020-neural} showed that transfer of the embedding layer is only beneficial if there is an overlap between the parent and child vocabulary such that embeddings for identical subwords can be aligned. Such alignments are very rare if the child language uses an unseen script.

\citet{gheini2019universal} train a universal vocabulary on multiple languages by romanizing languages written in a non-Latin script. Their many-to-one parent model can be transferred to new source languages without exchanging the vocabulary. In our work, we extend this idea to many-to-many translation settings using subsequent deromanization of the output. We study the trade-off between a greater vocabulary overlap and information loss as a result of romanization. Based on experiments on a diverse set of low-resource languages, we show that romanization is helpful for model transfer to related languages with different scripts.

\section{Romanization}
\label{sec:romanization}
Romanization describes the process of mapping characters in various scripts to Latin script. This mapping is not always reversible. The goal is to approximate the pronunciation of the text in the original script. However, depending on the romanization tool, more or less information encoded in the original script is lost. We compare two tools for mapping our translation input to  Latin script: 

\texttt{uroman}\footnote{\url{https://github.com/isi-nlp/uroman}} \citep{hermjakob-etal-2018-box} is a tool for universal romanization that can romanize almost all character sets. It is unidirectional; mappings from Latin script back to other scripts are not available. \texttt{uconv}\footnote{\url{https://linux.die.net/man/1/uconv}} is a command-line tool similar to \texttt{iconv} that can be used for transliteration. It preserves more information from the original script, which is expressed with diacritics. \texttt{uconv} is bi-directional for a limited number of script pairs.

Below is an example of the same Chinese sentence romanized with \texttt{uroman} and \texttt{uconv}:
\vspace{0.2cm}

\begin{center}
\begin{tabular}{c}
\begin{CJK*}{UTF8}{gbsn}
她到塔皓湖去了
\end{CJK*} \\
\texttt{uroman}: ta dao ta hao hu qu le\\
\texttt{uconv}: tā dào tǎ hào hú qù le\\
``She went to Lake Tahoe.''
\vspace{0.2cm}
\end{tabular}
\end{center}

The two tools exhibit different degrees of information loss. \texttt{uroman} ignores tonal information and consequently collapses the representations of \begin{CJK*}{UTF8}{gbsn}
塔
\end{CJK*} (Pinyin \textit{tǎ}; `tower') and \begin{CJK*}{UTF8}{gbsn}
她
\end{CJK*} (Pinyin \textit{tā}; `she'). Romanization with \texttt{uconv} retains this distinction but it still adds ambiguity and loses the distinction between \begin{CJK*}{UTF8}{gbsn}
她
\end{CJK*} (Pinyin \textit{tā}; `she') and \begin{CJK*}{UTF8}{gbsn}
他
\end{CJK*} (Pinyin \textit{tā}; `he'), among others. While \texttt{uconv} exhibits less information loss, its use of diacritics limits subword sharing between languages. We measure character-level overlap between English and romanized Arabic, Russian and Chinese with chrF scores \citep{popovic-2015-chrf} and find they are much higher for \texttt{uroman} (9.6, 18.8 and 13.3) compared to \texttt{uconv} (6.8, 18.1 and 7.2 respectively).

\section{Deromanization}
Romanization is not necessarily reversible with simple rules due to information loss. Therefore, previous work on romanized machine translation has focused on source-side romanization only \citep{DBLP:conf/aics/DuW17, wang-etal-2018-xmu, aqlan2019,briakou-carpuat-2019-university, gheini2019universal}. We argue that romanization can also be applied on the target side, followed by an additional deromanization step. This step can be performed by a character-based Transformer \citep{NIPS2017_7181} that takes data romanized with \texttt{uroman} or \texttt{uconv} as input and is trained to map it back to the original script. We provide more details on our deromanization systems in Appendix \ref{app:derom}. 

\section{Experimental Setup}

\subsection{Data}
We use OPUS-100 \citep{zhang-etal-2020-improving}\footnote{\url{https://github.com/EdinburghNLP/opus-100-corpus}}, an English-centric dataset that includes parallel data for 100 languages. It provides up to 1 million sentence pairs for every X-EN language pair as well as 2,000 sentence pairs for development and testing each. There is no overlap between any of the data splits across any of the languages, i.e. every English sentence occurs only once.

\begin{table*}
\centering
\begin{tabular}{cllcr}
&   & script    & related to & \# sentence pairs  \\ \midrule
\multirow{3}{*}{pretraining} & Arabic (ar) & Arabic     & he, mt    & 1,000,000  \\
\multirow{3}{*}{languages} & German (de) & Latin     & yi  & 1,000,000    \\
& French (fr) & Latin    & -   & 1,000,000   \\
 & Russian (ru) & Cyrillic    & sh  & 1,000,000  \\
& Chinese (zh) & Simpl. Han     & -   & 1,000,000  \\ \midrule
\multirow{3}{*}{(a)} & Amharic (am) & Ge'ez    & -  & 71,222  \\
& Marathi (mr) & Devanagari     & -    & 21,985  \\
& Tamil (ta) & Tamil    & -   & 198,927  \\ \midrule
\multirow{2}{*}{(b)} & Hebrew (he)* & Hebrew     & ar  & 50,000   \\
& Yiddish (yi) & Hebrew     & de & 7,718 \\ \midrule
\multirow{2}{*}{(c)} & Maltese (mt)* & Latin   & ar  & 100,000  \\
& Serbo-Croatian (sh)* & Latin    & ru     & 98,421
  
\end{tabular}
\caption{Overview of all languages, the script they are written in, other languages in this set they are closely related to (considering lexical similarity) and number of X$\leftrightarrow$EN sentence pairs. (*) means artificial low-resource settings were created.}
\label{tab:data}
\end{table*}

We pretrain our multilingual models on 5 high-resource languages that cover a range of different scripts \{AR, DE, FR, RU, ZH\} $\leftrightarrow$ EN. For our transfer learning experiments, we choose 7 additional languages that are either: \vspace{0.1cm}
\begin{itemize}
    \item[(a)] \textbf{Not closely related} to any of the pretraining languages and written in an \textbf{unseen script}, e.g. Marathi is not related to any of our pretraining languages and written in Devanagari script. \vspace{0.1cm}
    \item[(b)] \textbf{Closely related} to a pretraining language and written in an \textbf{unseen script}, e.g. Yiddish is related to German and written in Hebrew script.  \vspace{0.1cm}
    \item[(c)] Written in \textbf{Latin script} but \textbf{closely related to a pretraining language in non-Latin script}, e.g. Maltese is related to Arabic and written in Latin script.
\end{itemize} 
Our selection of low-resource languages covers a wide range of language families and training data sizes. Table \ref{tab:data} gives an overview of the selected languages.

\subsection{Model Descriptions}
We use \texttt{nematus}\footnote{\url{https://github.com/EdinburghNLP/nematus}} \citep{sennrich-EtAl:2017:EACLDemo} to train our models and SacreBLEU\footnote{BLEU+case.mixed+lang.XX-XX+numrefs.1\\+smooth.exp+tok.13a+version.1.4.2} \citep{post-2018-call} to evaluate them. We compute statistical significance with paired bootstrap resampling \citep{koehn-2004-statistical} using a significance level of 0.05 (sampling 1,000 times with replacement from our 2,000 test sentences). Our subword vocabularies are computed with byte pair encoding \citep{sennrich-etal-2016-neural} using the  SentencePiece implementation \citep{kudo-richardson-2018-sentencepiece}. We use a character coverage of 0.9995 to ensure the resulting models do not consist of mostly single characters.

\textbf{Bilingual Baselines:} We follow the recommended setup for low-resource translation in \citet{sennrich-zhang-2019-revisiting} to train our bilingual baselines for the low-resource pairs (original script). For our bilingual low-resource models, we use language-specific vocabularies of size 2,000.

\textbf{Pretrained multilingual models:} We pretrain three multilingual standard Transformer Base machine translation models \citep{NIPS2017_7181}: One keeps the original, non-Latin script for Arabic, Russian and Chinese (orig). The others (\texttt{uroman} and \texttt{uconv})  apply the respective romanization to these parent languages. We follow \citet{johnson-etal-2017-googles} for multilingual training by prepending a target language indicator token to the source input.  For our pretrained models, we use a shared vocabulary of size 32,000. An overview of our model hyperparameters is given in Appendix \ref{app:multi}.

\textbf{Finetuning:} We finetune our pretrained models independently for every low-resource language X. For finetuning on a child X$\leftrightarrow$EN pair, we use the same preprocessing as for the respective parent, i.e. we keep original script, use \texttt{uroman}, or use \texttt{uconv} for romanization. We reuse 250,000 sentence pairs from the original pretraining data and oversample the X$\leftrightarrow$EN data for a total of around 650,000 parallel sentences for finetuning. This corresponds roughly to a 3:2 ratio which helps to prevent overfitting. We early stop on the respective X$\leftrightarrow$EN  development set. For finetuning, we use a constant learning rate of 0.001. The remaining hyperparameters are identical to pretraining.

\subsection{Vocabulary Transfer} 
For our transfer baseline without romanization, we merge our bilingual baseline vocabulary with that of the parent model following previous work \citep{aji-etal-2020-neural, kocmi-bojar-2020-efficiently}. First, we align subwords that occur in both vocabularies. Next, we assign the remaining subwords from the bilingual baseline vocabulary to random unused positions in the parent vocabulary. With \texttt{uroman}, we can reuse the parent vocabulary as is. \texttt{uconv}, however, may produce unseen diacritics, which can result in a small number of unseen subwords. If that is the case, we perform the same vocabulary replacement for these subwords as for the vocabulary with the original script.

\section{Results}

\subsection{Does Romanization Hurt Translation?}
To study the effects of information loss from romanization, we compare the translation quality of our three pretrained multilingual models. To minimize the impact of deromanization, we only discuss X$\rightarrow$EN directions for languages with non-Latin scripts. The results are presented in Table \ref{tab:infoloss}. Whether romanization hurts the translation quality depends largely on the language pair. For example, for ZH$\rightarrow$EN, both romanization tools perform worse than the model trained on original scripts. This is in line with our previous discussion: Even though \texttt{uconv} keeps tonal information, there is still more ambiguity compared to using Chinese characters. The model trained with \texttt{uconv} romanization consistently outperforms \texttt{uroman}. This indicates that it is more important to minimize information loss than to maximize subword sharing.

\begin{table}
\centering
\begin{tabular}{cccc}
      & orig & uroman & uconv \\ \midrule
ar-en & \textbf{37.4} & 36.3   & \textbf{37.4} \\
ru-en & 33.3 & 33.5   & \textbf{34.1}  \\
zh-en & \textbf{39.5} & 37.0     & \textbf{39.2} 
\end{tabular}
\caption{X$\rightarrow$EN BLEU scores  \citep{papineni-etal-2002-bleu} of the multilingual pretrained models trained on original scripts (orig), romanized with \texttt{uroman} and \texttt{uconv}. Best systems (no other being statistically significantly better) marked in bold.}
\label{tab:infoloss}
\end{table}

An additional effect of using romanization, and thus being able to reuse the subword segmentation model during transfer, is that compression rates are worse than for dedicated segmentation models (see Table \ref{tab:lenghts}).
The resulting longer sequences with potentially suboptimal subword splits may also have a negative influence on translation quality.

\begin{table}
\centering
\begin{tabular}{rrrr}
   & orig    & uroman (\%)           & uconv (\%)    \\ \midrule
ar & 67.7     & + 2.2    & + 9.9      \\
de & 97.8     & - 0.5    & - 0.8      \\
fr & 131.7    & - 0.4   & - 0.6      \\
ru & 91.5     & + 3.3    & - 0.2      \\
zh & 54.1     & + 98.9   & + 156.6      \\ \midrule
am & 113.0    & + 70.4   & + 83.1      \\
he & 40.3     & + 17.6    & + 20.1      \\
mr & 42.4     & + 36.8    & + 35.4      \\
mt & 176.5    & - 1.9   & - 1.4      \\
sh & 168.0    & - 4.7   & - 5.5      \\
ta & 138.3    & + 20.1   & + 22.3      \\
yi & 54.2     & + 12.4    & + 39.5     
\end{tabular}
\caption{Average number of subwords per sentence with original script data (orig) and \% relative change after romanization (\texttt{uroman} and \texttt{uconv}). Original script data is segmented with a shared subword segmentation model for \{AR,DE,EN,FR,RU,ZH\} and  language-specific models for low-resource languages. For \texttt{uroman} and \texttt{uconv}, all languages are segmented using a shared model for \{AR,DE,EN,FR,RU,ZH\}, romanized with the respective tool.}
\label{tab:lenghts}
\end{table}

\subsection{Can We Restore the Original Script?}
Table \ref{tab:derom} compares our character-based Transformers to \texttt{uconv}'s built-in, rule-based deromanization. Relying on \texttt{uconv}'s built-in deromanization is not optimal. First, it does not support mappings back into all scripts. Second, the performance of built-in \texttt{uconv} deromanization
varies with the amount of ``script code-switching'', e.g. due to hyperlinks or email addresses. 
Character-based Transformers can learn to handle mixed script and outperform \texttt{uconv}'s built-in deromanization. 

Our models can reconstruct the original script much better from \texttt{uconv} data than from \texttt{uroman}. This is not surprising considering that \texttt{uroman} causes more information loss and ambiguity. As a shallow measure of the ambiguity introduced, we can compare the vocabulary size (before subword segmentation): With romanization, the total number of types in our training sets decreases on average by ~10\% for \texttt{uconv} and by ~14\% for \texttt{uroman}. 

Preliminary experiments with artificial low-resource settings (Appendix \ref{app:derom_sizes}) showed that additional training data can improve deromanization but it performs well even with very small amounts of training data (10,000 sentences). This shows that our proposed character-based Transformer models are powerful enough to learn a mapping back to the original script as much as this is possible, given the increased ambiguity. This finding is supported by concurrent work showing that character-based Transformers are well-suited to a range of string transduction tasks \citep{wu2020applying}.

\begin{table}
\centering
\begin{tabular}{cccc}
                        & built-in & \multicolumn{2}{c}{learned} \\ \cmidrule(lr){2-2}\cmidrule(lr){3-4}
\multicolumn{1}{c}{}   & uconv    & uconv            & uroman    \\ \midrule
\multicolumn{1}{c}{ar} & 92.7     & \textbf{98.1}    & 94.6      \\
\multicolumn{1}{c}{ru} & 94.9     & \textbf{99.2}    & 98.8      \\
\multicolumn{1}{c}{zh} & -        & \textbf{96.7}    & 94.0      \\ \midrule
\multicolumn{1}{c}{am} & -        & \textbf{99.7}    & 97.8      \\
\multicolumn{1}{c}{he} & 98.7     & \textbf{99.7}    & 96.9      \\
\multicolumn{1}{c}{mr} & 79.1     & \textbf{99.3}    & 97.6      \\
\multicolumn{1}{c}{ta} & 72.9     & \textbf{98.3}    & 98.0      \\
\multicolumn{1}{c}{yi} & 49.6     & \textbf{96.9}    & 89.6     
\end{tabular}
\caption{chrF scores of the deromanization to the original script. Best systems marked in bold.}
\label{tab:derom}
\end{table}

\subsection{Transfer to Low-Resource Languages}
\label{sec:transfer}
Table \ref{tab:results} shows the results from our experiments on transfer learning with romanization. Romanizing non-Latin scripts is not always useful. For low-resource languages that use an unseen script but are not related to any of the pretraining languages (a), the performance degrades for \texttt{uroman} and is not statistically significantly different for \texttt{uconv}. The extremely low BLEU score for EN$\rightarrow$AM shows another problem with \texttt{uroman} romanization: \texttt{uroman} ignores the Ethiopic word space character which increases the distance between translation and reference.

However, for languages that are related to a pretraining language with a different script (groups (b) and (c)), there is an added benefit of using romanization. The statistically significant improvement of \texttt{uconv} over \texttt{uroman} strengthens our claim that it is important to keep as much information as possible from the original script when mapping to Latin script. Despite potential information loss from romanization and error propagation from deromanization, our results show that romanization has merit when applied to related languages that can profit from a greater vocabulary overlap.

\begin{table}[ht]
\centering
\begin{tabular}{lrrrrr}
&&& \multicolumn{3}{c}{transfer from}\\ 
&&& \multicolumn{3}{c}{multilingual parent}\\ \cmidrule(r){4-6} 
                                          & \multicolumn{1}{r}{} & \multicolumn{1}{r}{base} & \multicolumn{1}{r}{orig} & \multicolumn{1}{r}{uroman} & \multicolumn{1}{r}{uconv} \\ \midrule
\multirow{7}{*}{(a)} & am-en                 & 14.4                             & \textbf{16.2}                                 & \textbf{16.5}               & \textbf{16.0}                      \\
                     & en-am                 & 12.7                             & 13.7                                 & \phantom{0}6.5                         & \textbf{14.3}             \\
                     & mr-en                 & 34.3                             & \textbf{45.0}                          & 43.4                        & 42.8                      \\
                    & en-mr                 & 25.7                             & \textbf{33.4}                        & \textbf{33.2}                        & \textbf{33.0}                      \\
                     & ta-en                 & 21.9                             & \textbf{29.3}                     & \textbf{29.0}                        & \textbf{29.2}                      \\
                     & en-ta                 & 13.5                             & \textbf{21.5}                                 & 21.0                        & \textbf{22.4}             \\ \cmidrule(r){2-6} 
                    & avg imp                  & -                             & \textbf{+ 6.1}                                 & + 4.5                        & + \textbf{5.9}             \\ \midrule
\multirow{5}{*}{(b)} & yi-en                 & \phantom{0}6.9                              & 22.5                                 & 24.9                        & \textbf{28.9}             \\
                     & en-yi                 & \phantom{0}9.5                              & 12.0                                 & \textbf{20.7}               & \textbf{19.7}                     \\
                    & he-en                 & 22.8                             & \textbf{28.6}                                 & \textbf{28.5}                        & \textbf{29.0}             \\
                   & en-he                 & 21.1                                 & 24.5                                & 25.2                            & \textbf{26.6}                 \\ \cmidrule(r){2-6} 
                     & avg imp                  & -                             & + 6.8                                 & + 9.8                       & \textbf{+ 11.0}         \\ \midrule
\multirow{5}{*}{(c)} & mt-en                 & 46.5                             & \textbf{59.1}                                 & \textbf{59.5}               & \textbf{59.5}             \\
                     & en-mt                 & 35.6                             & \textbf{45.0}                                   & \textbf{45.2}                        & \textbf{45.3}             \\
                    & sh-en                 & 40.1                                 & 55.5                                 & \textbf{56.3}                        & \textbf{56.7}             \\
                    & en-sh                 & 33.8                                 & 52.1                                 & 52.3                        & \textbf{53.7}    \\ \cmidrule(r){2-6} 
                    & avg imp                  & -                             & + 13.9                                 & + 14.3                       &  \textbf{+ 14.8}                    
\end{tabular}
\caption{BLEU scores of the bilingual baselines (no transfer learning) and finetuned models using original scripts (orig), romanized with \texttt{uroman} and \texttt{uconv}. Average improvement over bilingual baseline is shown per group of languages. Best systems (no other being statistically significantly better) marked in bold.}
\label{tab:results}
\end{table}
\section{Conclusion}
We analyzed the value of romanization for transferring multilingual models to low-resource languages with different scripts.
While we cannot recommend romanization as the default strategy for multilingual models and transfer learning across scripts because of the information loss inherent to it, we find that it benefits transfer between related languages that use different scripts.
The \texttt{uconv} romanization tool outperforms \texttt{uroman} because it preserves more information encoded in the original script and consequently causes less information loss.
Furthermore, we demonstrated that romanization can also be successful on the target side if followed by an additional, learned deromanization step. We hope that our results provide valuable insights for future work in transfer learning and practical applications for low-resource languages with unseen scripts.

\section*{Acknowledgements}
We thank our colleagues Anne, Annette, Duygu, Jannis, Mathias, Noëmi
and the anonymous reviewers for their helpful feedback. This work was funded by the Swiss National Science Foundation (project MUTAMUR;
no. 176727).

\bibliography{anthology,emnlp2020}
\bibliographystyle{acl_natbib}
\newpage
\appendix
\section{Model Details}
\subsection{Multilingual Pretrained Models}
\label{app:multi}
We train multilingual Transformer Base machine translation models \citep{NIPS2017_7181} with 6 encoder layers, 6
decoder layers, 8 heads, an embedding and hidden state dimension of
512 and a feed-forward network dimension of 2048. We regularize our models with a dropout of 0.1 for the embeddings, the residual connections, in the feed-forward sub-layers and for the attention weights. Furthermore, we apply exponential smoothing of 0.0001 and label smoothing of 0.1. We tie both our encoder and decoder input embeddings as well as the decoder input and output embeddings \citep{press-wolf-2017-using}. All of our multilingual machine translation models are trained with a maximum token length of 200 and a vocabulary of size 32,000.

For optimization, we use Adam \citep{DBLP:journals/corr/KingmaB14} with standard hyperparameters and a learning rate of  0.0001. We follow the Transformer learning schedule described in \citep{NIPS2017_7181} with a linear warmup over 4,000 steps. Our token batch size is set to 16,348 and we train on 4 NVIDIA Tesla V100 GPUs. All models were trained using the implementation provided in \texttt{nematus} \citep{sennrich-EtAl:2017:EACLDemo} using early stopping on a development set with patience 5.

\subsection{Character-Based Deromanization}
\label{app:derom}
We train character-based Transformer Base machine translation models \citep{NIPS2017_7181}. To achieve character-level deromanization, we do not make any changes to the architecture. We simply change the input format such that every character is separated by spaces. The original space characters are replaced by another character that does not occur in the training data ($\diameter$). The following example shows the parallel training data for learned deromanization:
\vspace{-0.2cm}
\begin{center}
\begin{tabular}{c}
\texttt{uroman} source: C H t o $\diameter$ t a m $\diameter$ d a l s h e ?\\
\texttt{uconv} source: Č t o $\diameter$ t a m $\diameter$ d a l ' š e ?\\
target: \foreignlanguage{russian}{Ч т о $\diameter$  т а м $\diameter$ д а л ь ш е ?}  \\
``What's next?''
\vspace{0.2cm}
\end{tabular}
\end{center}

We use a maximum sequence length of 1,200 since character-level sequences are much longer than subword-level sequences. Our vocabularies are made up of all characters that occur in the respective training data. All other parameters are set as for multilingual pretraining described in Appendix \ref{app:multi}.

\section{Supplementary Results}
\subsection{Effect of Data Size on Deromanization}
\label{app:derom_sizes}
\begin{figure}[ht]
    \centering
    \includegraphics[width=0.4\textwidth]{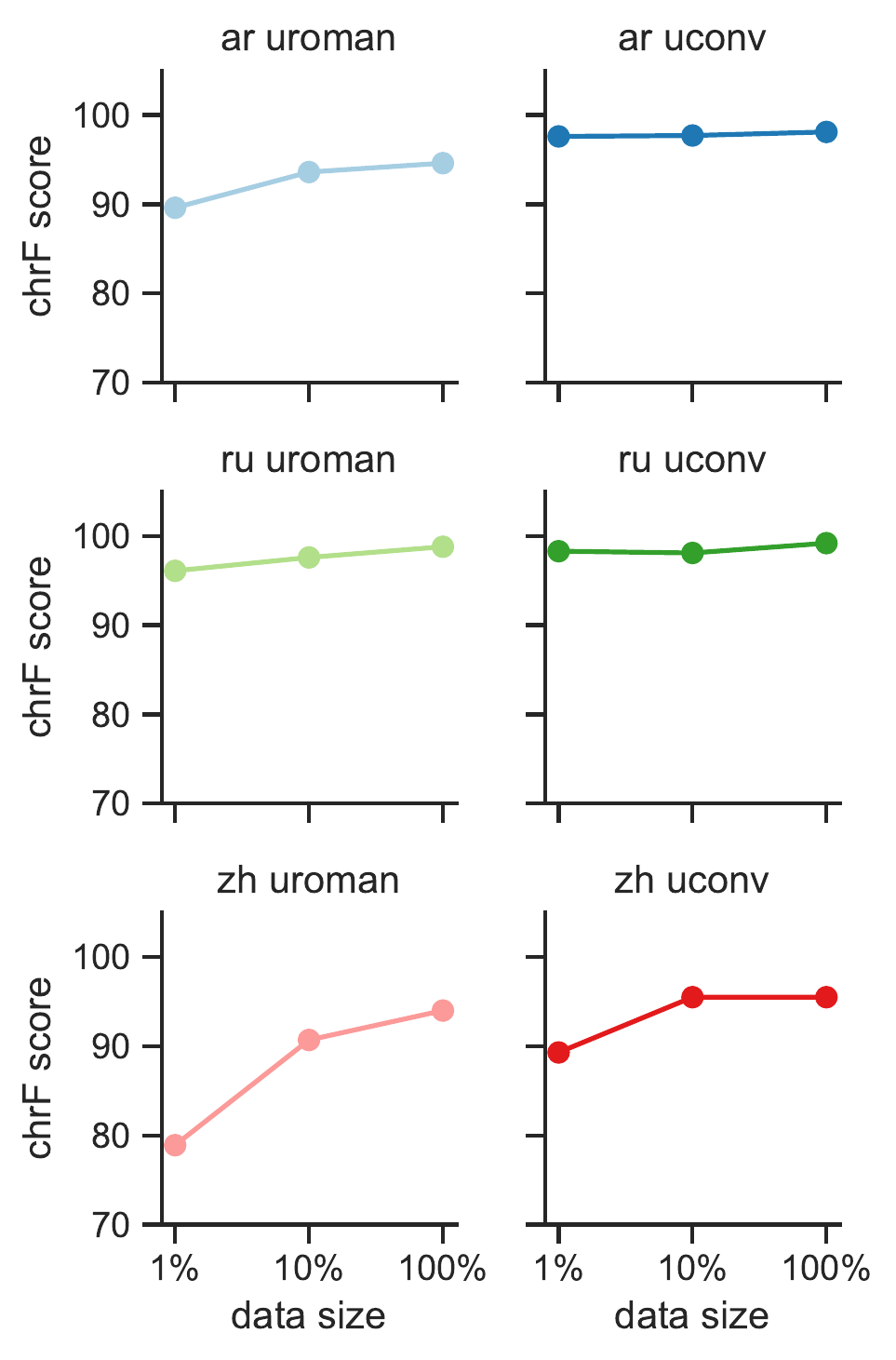}
    \caption{chrF scores of deromanization models trained on 1\%, 10\% and 100\% of the total data (corresponding to 10,000, 100,000 and 1,000,000 parallel sentences). Results compare romanization with \texttt{uroman} and \texttt{uconv} for Arabic, Russian and Chinese.}
    \label{fig:derom_graph}
\end{figure}

\begin{table}[ht]
\centering
\begin{tabular}{lccc}
      & \multicolumn{3}{c}{uroman}  \\ \cmidrule(r){2-4} 
      & 1\% & 10\% & 100\%  \\ \midrule
en-ar & 20.3 & \textbf{21.6}  & \textbf{21.7}   \\
en-ru & 26.5 & 27.9   & \textbf{28.5} \\
en-zh & 38.8 & \textbf{41.6}   & \textbf{41.9} \\ \vspace{-0.2cm}
\end{tabular} 
\begin{tabular}{lccc}
  
      &  \multicolumn{3}{c}{uconv} \\  \cmidrule(r){2-4}
      & 1\% & 10\% & 100\% \\ \midrule
en-ar & 21.2 & \textbf{21.8} & \textbf{21.7}  \\
en-ru & 27.9 & 28.2 & \textbf{29.3}\\
en-zh & 40.2 & \textbf{41.8} & \textbf{41.9}
\end{tabular}
\caption{EN$\rightarrow$X BLEU scores of the multilingual pretrained models after deromanization. Deromanization models were trained on 1\%, 10\% and 100\% of the total data (corresponding to 10,000, 100,000 and 1,000,000). Best systems (no other being statistically significantly better)  marked in bold.}
\label{tab:bleuderom}
\end{table}

Figure \ref{fig:derom_graph} shows the influence of the training data size on the chrF score between the deromanized test set and the original script test set. Additional data can improve deromanization models, especially for languages such as Chinese, where a mapping back to the original script is difficult to learn due to the information loss from romanization. 

We analyze how deromanization quality affects the BLEU score of deromanized translations. This is shown in Table \ref{tab:bleuderom}. We find that the deromanization models for \texttt{uroman} are more affected by an extreme low-resource setting. For \texttt{uconv}, deromanization models trained on smaller data sets show less performance loss compared to using full data. It is notable that training \texttt{uconv} deromanization models only on 100,000 sentences has almost no effect on the BLEU score for EN$\rightarrow$AR and EN$\rightarrow$ZH. For EN$\rightarrow$RU, there is a loss of 1.1 BLEU points compared to training on 100\% of the data. Looking at the deromanization outputs for EN$\rightarrow$RU, we found that deromanization models trained on less data could not handle ``script code-switching'' as well as the models trained on full data. 

While these results show that additional training material can improve deromanization, they do not mean that romanization on the target side cannot be used in low-resource machine translation settings. First, our results in Section \ref{sec:transfer} have shown that romanization on the target side can bring improvements even if deromanization models cannot perfectly reconstruct the original script. Second, it will often be possible to find additional monolingual data to improve deromanization models.

\end{document}